# Experiments with truth using Machine Learning: Spectral analysis and explainable classification of synthetic, false, and genuine information

Vishnu S. Pendyala
Department of Applied Data Science
San Jose State University
San Jose, CA 95192-0250
Email: vishnu.pendyala@sjsu.edu

Madhulika Dutta
Department of Applied Data Science
San Jose State University
San Jose, CA 95192-0250
Email: madhulika.dutta@sjsu.edu

*Abstract*—Misinformation is still a major societal problem and the arrival of Large Language Models (LLMs) only added to it. This paper analyzes synthetic, false, and genuine information in the form of text from spectral analysis, visualization, and explainability perspectives to find the answer to why the problem is still unsolved despite multiple years of research and a plethora of solutions in the literature. Various embedding techniques on multiple datasets are used to represent information for the purpose. The diverse spectral and non-spectral methods used on these embeddings include t-distributed Stochastic Neighbor Embedding (t-SNE), Principal Component Analysis (PCA), and Variational Autoencoders (VAEs). Classification is done using multiple machine learning algorithms. Local Interpretable Model-Agnostic Explanations (LIME), SHapley Additive exPlanations (SHAP), and Integrated Gradients are used for the explanation of the classification. The analysis and the explanations generated show that misinformation is quite closely intertwined with genuine information and the machine learning algorithms are not as effective in separating the two despite the claims in the literature.

## I. INTRODUCTION

Misinformation, fake news, and lies have been adversely impacting society in many significant ways. The advent of generative AI applications such as ChatGPT has only exasperated the problem [1]. Misinformation can be multi-modal. Generative AI is capable of producing misinformation in multiple modalities. However, while the difficulties in detecting fake images are well documented in the literature [2], the same is not true for misinformation in the form of text. On the other hand, some studies have reported 100% accuracy in detecting AI-generated text using simple language models such as BOW [3]. There is also abundant literature on solving the misinformation containment problem with human-generated text but it is a well-known fact that the problem is still largely unsolved [4]. This work expands on the previous analysis of the problem [5].

Given the current gap in the literature in sufficiently identifying the reasons why misinformation containment is still an unsolved problem, there is a need to focus on why machine learning is unable to solve the problem despite the tall claims to the contrary in the literature, which is mostly based on the currently available embedding techniques. Embeddings are essentially representations of the input data in the hidden layers of neural networks. This work is an attempt to determine what makes it so difficult to identify misinformation based on the embeddings and the limitations of the current Neural Natural Language Processing (NNLP) and Machine Learning (ML) techniques in doing so using a variety of experiments to visualize, classify, and explain.

## II. RELATED WORK

Misinformation containment is proven in the literature to be NP-hard [6]. Misinformation detection can be addressed using diverse approaches, including algorithms such as the Kalman Filter [7], statistical techniques and first-order logic [8]. However, it is established in the literature that machine learning is a good alternative to heuristic algorithms to solve NP-hard problems [9]. A further literature survey naturally shows a comprehensive use of machine learning and deep learning in conjunction with NLP techniques to address the problem. There are multiple surveys [10] [11] describing the literature in this regard. Large language models (LLMs) have also been used to detect misinformation and the reasoning behind the classifications has been discussed qualitatively and quantitatively using explainability techniques [12].

The explainability perspective of misinformation detection has also been explored in the literature. Liu et al. [13] propose a logic-based neural model for multimodal misinformation detection, integrating interpretable logic clauses to enhance explainability and reliability in detecting misinformation on online social platforms. In another research [14], a method combining two approaches is proposed: (a) Domain Adversarial Neural Network (DANN) for generalizability across platforms and (b) an explainable AI technique, LIME to understand the model's reasoning. The authors test this method on COVID-19 misinformation and show that it significantly improves detection accuracy while providing explanations for the results. Explainable AI and visualization techniques to enhance understanding of disinformation detection models are proposed to aid in interpreting and presenting results effec-tively [15]. A framework called DISCO [16] for disinformation detection provides explainability by identifying misleading words and utilizing a model-agnostic feature extraction scheme for transparency. Literature review shows that more papers [17] use explainability techniques to interpret misinformation detection.

### A. Contribution

Most of the current literature detects misinformation using prevalent embedding frameworks. This paper investigates the following research question: *RQ: How effective are the current embedding techniques in separating truthful information from false information?* The answer to the question is approached from visualization first followed by generating insights into how the model arrived at the classification. The work is unique in framing the research question and investigating it from diverse perspectives. In the literature survey, a thorough quantitative analysis of the determinants found that the homogeneity of the communities in terms of their information consumption pattern is the primary driver for misinformation spread [18]. Other than that, to the best of our knowledge, this is the first work that investigates using spectral and non-spectral methods of visualization and explainability, why misinformation containment is mostly an unsolved problem using diverse datasets containing synthetic, genuine, and false information.

## III. DATASETS AND PREPROCESSING

For a comprehensive analysis, a variety of datasets are used for the investigation.

### A. LIAR Dataset

The LIAR dataset [19] contains around 12,800 short statements collected from various sources such as political debates, Facebook posts, news releases, and tweets that are labeled manually. There are six fine-grained labels for each of the 12,800 statements: true, mostly-true, half-true, barely-true, false, and pants-fire, indicating the degrees of truthfulness. For this project, the labels were encoded into three categories, the first class consists of all true labels, the second class contains false, and the third class contains pants-fire statements. After encoding the labels, the dataset has 1047 'pants-fire' statements, 2507 'false' statements, and 9237 'true' statements. After splitting the data into train, test, and validation sets, the true and false statements in the train set were downsampled to 839 statements in each of the three classes to get a balanced dataset. Since language models such as BERT are pre-trained, not having a huge dataset is not an issue. The usual preprocessing of the dataset was done to make it ready for the classification task. For instance, to rule out any non-English occurrences in the statements, all statements containing non-ASCII characters were removed.

### B. Human ChatGPT Comparison Corpus

The dataset used for the second set of experiments is a Question Answering Dataset based on the public Human ChatGPT Comparison Corpus (HC3) data. This dataset's main objective is to compare and examine the variations between responses produced by ChatGPT and human respondents in a variety of disciplines, such as open-domain, finance, health, law, and psychology. The dataset contains various question prompts, and answers given by ChatGPT and humans. For the experiments, the questions that were asked from ChatGPT were not used. Only the answers given were used. The dataset contains 24,321 question-answer prompts, but for experimentation, only the first 15,000 answers from humans and ChatGPT were used. The dataset was already clean and balanced, hence pre-processing was not needed. To train the classification model, 15,000 answers each for both classes were selected. Then the data was shuffled and split into the train, test, and validation sets in the ratio of 8:1:1. The data was split in such a way that the train, test, and validation sets had an equal distribution of classes.

### C. Artificial Intelligence Generated Abstracts

The Artificial Intelligence Generated Abstracts (AI-GA) dataset sourced from GitHub consists of 28,662 entries for scientific paper titles, extracted from the COVID-19 Open Research Dataset Challenge (CORD-19) dataset, and their abstracts. For half of the paper titles, the abstracts were generated using GPT-3 and are thus not original. The dataset is prelabeled, with label 0 signifying real human-written abstracts and label 1 for artificially generated ones. Each class has 14,331 instances belonging to it, making the dataset balanced, and no additional pre-processing was needed to be performed.

## IV. METHODOLOGY

For carrying out spectral analysis of misinformation in text, the statements in the dataset are converted into vectors using embeddings techniques based on BERT [20], S-BERT [21], which build upon the transformer architecture [22] and also on the Doc2VecC framework [23]. The embeddings project the statements into the latent feature vector space. Spectral analysis and visualization of the statement vectors are then performed using two substantially different techniques - the t-SNE [24] and PCA [25] algorithms depicting the actual labels. PCA is a linear approach, while t-SNE is a non-linear visualization technique. Hence the two are fundamentally different in their approach. The diversity in the representation learning, classification algorithms, and spectral techniques is to ensure that as much as possible, no bias in our conclusions is attributable to the technique.

The dimensionality is reduced to two so that the latent feature space can be visualized better. The analysis shows that the data is highly non-linear. The statements are then classified using two machine learning algorithms, Support Vector Machines (SVM) using the Radial Basis Function (RBF) kernel and K-Nearest Neighbors (KNN), which are known to perform well on non-linear data. These classification algorithms work in fundamentally different ways as well. The hyperparameters such as the kernel function used for SVM and the value of k for KNN are determined through the process of validation, trying out several alternatives and choosing the best ones. PCA and t-SNE are run again on the data, but this time, depicting the predicted labels.

To further test the hypothesis that the current NLP embedding frameworks are inadequate and explore other techniques to differentiate human-generated and artificial text, we experimented with two other data sources. Answer statements from the training set of the Question Answering dataset were used to train the Robustly Optimized BERT Pretraining Approach (RoBERTa) [26] natural language model. The model's performance on the predictions on the test set was analyzed using known evaluation metrics. To interpret the model's reasoning behind the classification of the text as fake or original, explainability algorithms, Local Interpretable Model-Agnostic Explanations (LIME) [27], SHapley Additive exPlanations (SHAP) [28], and Integrated Gradients [29] were used on the output of the classifier.

In studying the AI-GA data source containing scientific paper titles and abstracts, the dataset was split into train and test sets in the ratio of 75:25. Text from the 'abstract' column of the train and test datasets is converted into sentence-level embeddings using the Sentence Transformer library. The model 'roberta-base-nli-stsb-mean-tokens' is used as it was found to be better suited for natural language processing tasks involving classification. The embeddings generated are input to the support vector machines algorithm to train a model and use it to classify the text as belonging to fake or original abstracts.

To find out any underlying patterns in the fake and original text, data visualization was done using variational autoencoders [30], a type of deep learning framework that uses a probability-based approach to model input sample data. The predicted instances were analyzed using LIME and SHAP libraries to improve the interpretability of the model output.

## V. EXPERIMENTS

### A. Spectral Analysis of Misinformation

Three experiments were conducted on the LIAR dataset to detect the latent patterns in the data. To be able to apply different methods to capture the pattern of true, false, or lie statements it was needed to represent the statements as vectors concerning their context. These experiments were done as follows:

*1) Statement Embeddings using BERT*

The pre-trained model called bert-base-uncased from the Huggingface open-source library is used to vectorize the statements. The dataset is downsampled to make it balanced. The final generated vectors are present in a 2D array with 2517 rows and 768 embedding dimensions. The results of PCA and t-SNE are visualized in Fig. 1 and Fig. 2. Both graphs are highly scattered, showing the features are highly non-linear and cannot be clustered in a 2D space.

*2) Statement Embeddings Using Doc2vecC*

As the t-SNE and PCA results of the embeddings using the BERT model showed no specific pattern, the doc2vecC model was applied to the LIAR dataset. The Doc2VecC framework

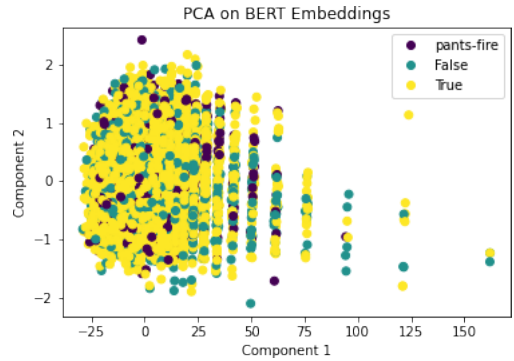

Fig. 1. Applying PCA to BERT Embeddings using mean of the word tokens

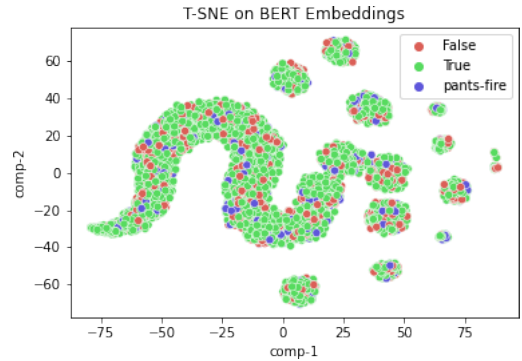

Fig. 2. Applying t-SNE on BERT Embeddings Using Mean of the word tokens

[23] represents a document as an average of the word embeddings of randomly sampled words in that document. Word embeddings generated by Doc2vecC with respect to the context are reported to be significantly better than those generated by Word2Vec [23]. An additional corruption model is included in the algorithm that gives more importance to informative words and suppresses common words by using data-dependent regularization. The original Doc2VecC algorithm produces a vector consisting of 100 dimensions. For this project, it is changed to 256 dimensions to capture more features from each statement. The visualizations using PCA and t-SNE are shown in Fig. 3 and Fig. 4. The resultant graphs are again highly scattered with no specific pattern.

*3) Statement Embeddings Using Sentence-BERT*

Sentence-BERT (SBERT) [21] is a modification of the pre-trained BERT algorithm. SBERT embeds sentences faster and more accurately in comparison to BERT and its optimized variant, RoBERTa. This model maps sentences and paragraphs to a 384-dimensional vector and can be used for tasks like clustering or semantic search. To vectorize the statements of the LIAR dataset, a pre-trained model "all-MiniLM-L6-v2" was applied. Post this, PCA and t-SNE were applied to the statement embeddings to visualize the dataset. The results of these two methods are shown in Fig. 5 and Fig. 6. The obtained graphs are highly scattered with no specific pattern.

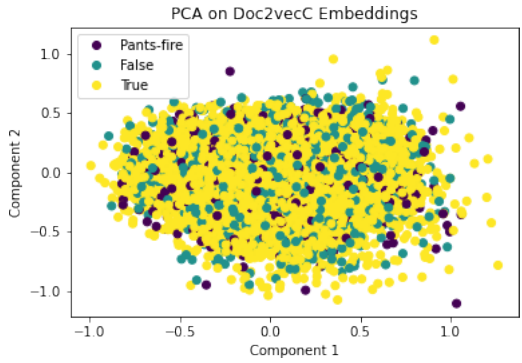

Fig. 3. Applying PCA to Doc2vecC Embeddings

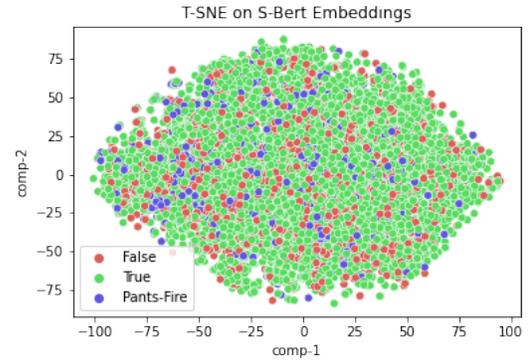

Fig. 6. Applying t-SNE on S-BERT Embeddings

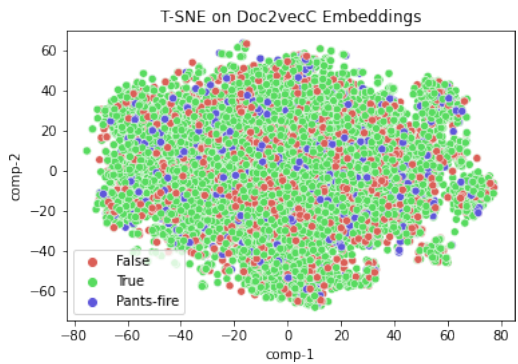

Fig. 4. Applying t-SNE on Doc2vecC Embeddings

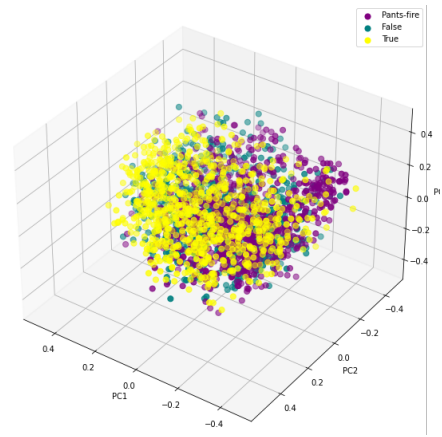

Fig. 7. Applying Three Dimensional PCA on S-BERT Embeddings predicted Labels Using SVM

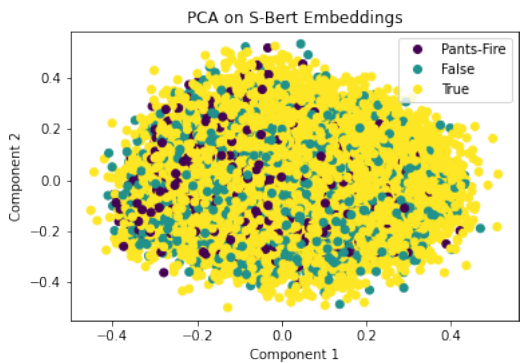

Fig. 5. Applying PCA to S-BERT Embeddings

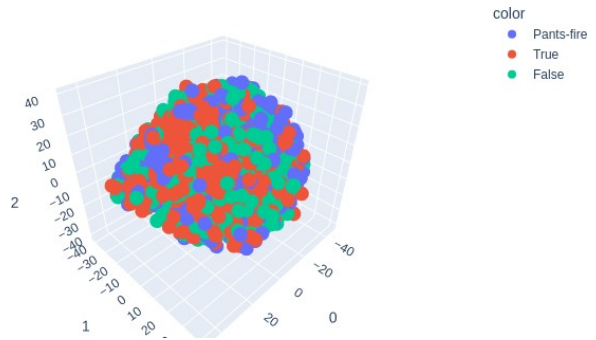

Fig. 8. Applying Three Dimensional t-SNE on S-BERT Embeddings predicted Labels Using SVM

Classification of Embeddings:

The vectors generated by BERT, Doc2vecC and SBERT were used for classification using K-Nearest Neighbors (KNN) and Support Vector Machine (SVM). Since this is a multi-class problem, the decision function for the SVM algorithm has been chosen to be "ovo," which stands for "one-vs-one". Based on our experiments during validation, for the KNN classifier, six nearest neighbors were considered (k=6), using the Minkowski distance metric, uniform weights, and kd-tree structure.

### B. AI/Human Generated Text Detection

To study different techniques that could help distinguish between text written by humans and synthetically generated text, the following experiments were carried out:

1) ChatGPT versus Human Answered Questions

To classify a text if it has been generated by ChatGPT or Human, the pre-trained "roberta-base" model from the transformers library was used. The RobertaForSequenceClas-

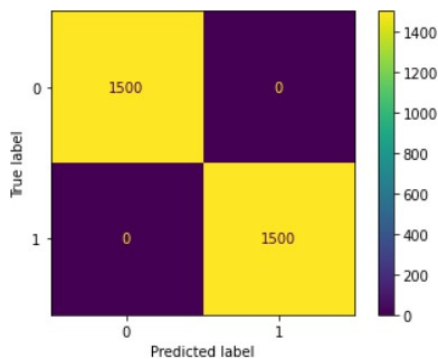

Fig. 9. Confusion Matrix for RoBERTa Classifier on HC3 Dataset

```
              precision    recall  f1-score   support

           0       1.00      1.00      1.00      1500
           1       1.00      1.00      1.00      1500

    accuracy                           1.00      3000
   macro avg       1.00      1.00      1.00      3000
weighted avg       1.00      1.00      1.00      3000
```

Fig. 10. Classification Report for RoBERTa Classifier on HC3 Dataset

sification class from the Transformers library was used to train our RoBERTa model. It is a RoBERTa Model transformer with a sequence classification head on top. The classification head is a linear layer on top of the pooled output. Upon analyzing the input data, it was found that the median length of a sentence in the input texts is 144, and the 75 percentile is at 199, so the max_length hyperparameter of the RoBERTa model was set to 256, and padding was enabled. The model was then fine-tuned using the PyTorch library for a total of 2 epochs, with a batch size of 32. AdamW from the huggingface library was used as the optimizer while fine-tuning.

After training, the model was tested on the test set resulting in an F1 score of 1.00. The confusion matrix and classification report obtained after performing the tests can be seen in Fig. 9 and Fig. 10 respectively. The training data for the model classification contains answers for prompts having "Explain like I'm a five-year-old" at the end of the query. The high F1 score may be due to the low diversity of Question-Answer prompts in the training and test data.

After the classification model was trained and ready for use, interpretation techniques such as the Integrated Gradients and LIME were used to try and interpret the results of the classifier. Captum library [31] was used to implement the Integrated Gradients technique and generate attribution scores to understand which features in the input text played an important role in determining whether the texts are generated by ChatGPT or Human.

2) Differentiate Real and Fake Scientific Text

With the advent of generative large language models, it could become increasingly common to encounter non-original text even in academic settings. To exemplify this we selected a dataset where half of the scientific abstracts are generated using the GPT-3 language model, while the other half are written by humans. All the sentences were converted from these abstracts into fixed-length embeddings using the roberta-base-nli-stsb-mean-tokens model from the Sentence Transformer library. This is done so that the text can be represented in a numeric format that is understood by machine learning classifier algorithms. We pass these embeddings to an SVM model built using the polynomial kernel and one-vs-one or 'ovo' density function. After training the model on embeddings for 21,496 abstracts, we test it on a corpus of 7,166 abstract entries. The metrics accuracy, precision, recall, and F1 score were calculated to evaluate the performance of the model.

The next step is to explore if there are any significant patterns in the distribution of the generated embeddings that would help distinguish one class of abstracts from the other. Variational autoencoders are a type of deep learning model that uses a probability-based approach to model sample input data. Using the TensorFlow library, an autoencoder model was created with relu activation function in the input encoder layer and sigmoid activation in the output decoder layer and fitted to the text of the abstracts. While compiling results from the autoencoder, the Mean Squared Error (MSE) and cosine embedding loss functions were used to observe differences in the generated samples. The resulting visualization models the data along its two most significant components identified and shows the distribution of the fake and original abstracts in this space. Fig. 11 shows the visualization using the loss based on cosine similarity, while Fig. 12 shows the one based on MSE.

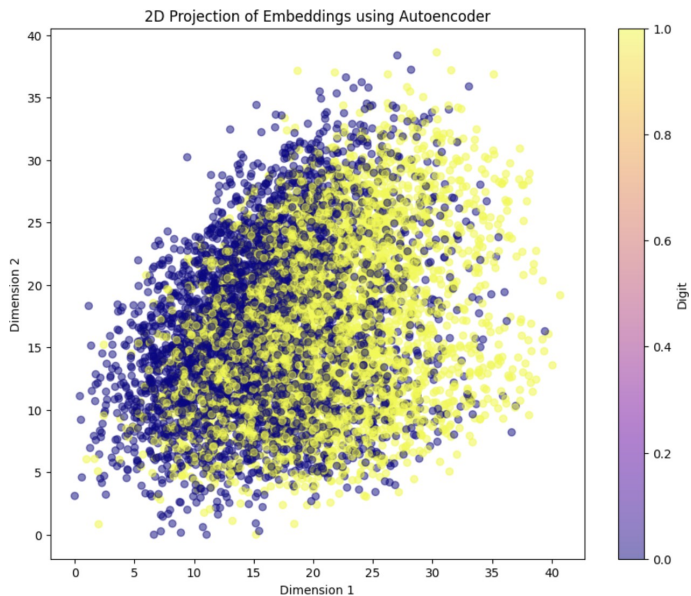

Fig. 11. Autoencoder Visualization of S-bert Embeddings using Cosine Similarity

As most machine learning algorithms are black-box models, it is often difficult to interpret the generated results for common end users to answer questions such as why a data item was classified as such. In an attempt to explain the ratio-nale behind the model's classification, the Local Interpretable Model-agnostic Explanations (LIME) and SHapley Additive exPlanations (SHAP) Python packages were used. The text from the abstract column of the dataset was converted to a vector representation using the CountVectorizer feature extraction technique. The vectors are passed through a logistic regression model this time and after fitting on the training dataset, the model was used to predict the class of the sentences in the test dataset. These two steps of vectorization and classification are combined using a pipeline function given as input to the LIME explainer package. Fig. 13 shows the in-text significant word identification using LIME. The features, in this case, words, that are most important for the model's class prediction are listed in the middle of the figure, along with their weights. The weight of a feature indicates how important that feature was for the model's prediction. For example, the word "study" has the most weight of 0.05, which means that it was the most important feature for the model's prediction. As can be seen, there are no significant words in the text that influence the model towards a '1' class prediction. All the words influence the model towards a '0' class prediction.

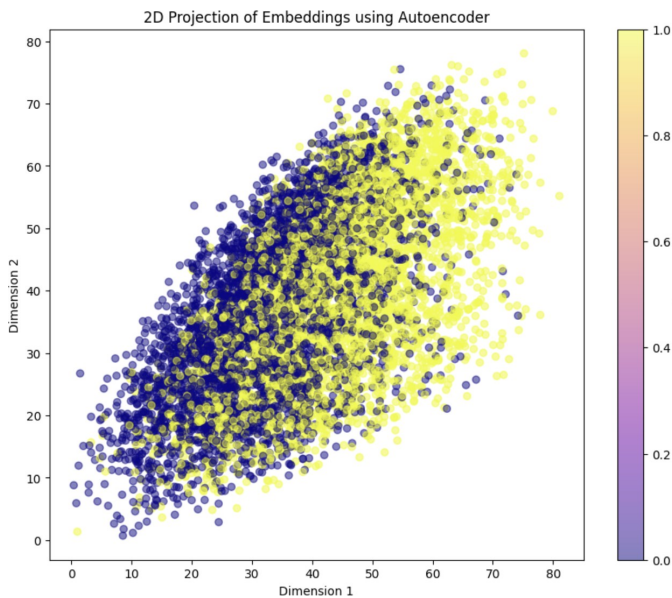

Fig. 12. Autoencoder Visualization of S-bert Embeddings using Mean Squared Error (MSE)

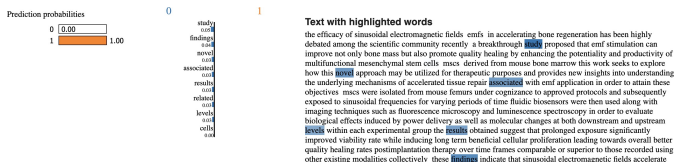

Fig. 13. Text Classification Model Explainability using LIME

To implement SHAP, the sentences are tokenized similarly using TensorFlow text preprocessing packages and passed to the explainer package. Outputs from both were compared to identify specific words in the sentences that could indicate them as being human-written or artificially generated. Fig. 14 is a summary plot from SHAP that shows the contribution of the top few words towards the prediction. The words are shown on the y-axis. The x-axis shows the SHAP value of each of these words. The magnitude of the value represents how much the feature impacts the prediction. The color represents the word's influence on the model towards a class. In this specific case, words such as "data," "control," and "conclusion" pull the classification towards 1, as denoted by the length of the blue bars. On the other hand, words such as "associated," "during," and "can" influence the model towards a 0 prediction based on the length of the red bars.

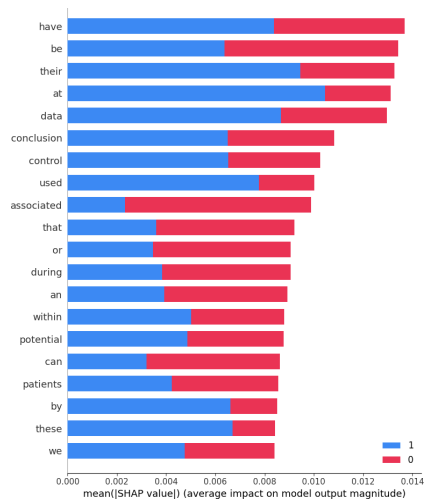

Fig. 14. In-text Significant Word Identification using SHAP

## VI. RESULTS

For comprehensiveness, the performance of the SVM and KNN models was evaluated using different metrics. Accuracy shows how well the models predicted the true-false and true-negatives. Precision emphasizes true positives and the F-1 score shows a harmonic mean of precision and recall. In addition to accuracy, F-1 score, and precision, ROC accuracy was calculated. In a multi-class classification experiment with a highly imbalanced dataset, ROC micro-averaging is preferable over macro-averaging. Micro-averaging will aggregate the contributions of all classes to calculate the average. However, in macro-averaging, the ROC is calculated independently of each class, and then an average is applied to it that causes it to treat each class equally. Thus, for the balanced dataset, it is preferable to use ROC macro-averaging.

The performance results of classifications using SVM and KNN models for BERT, doc2vecC, and SBERT embeddings are shown in Table I.

In Fig. 15 the ROC accuracy is shown for true vs the rest of the classes. The area under the curve gave a 65% accuracy but as shown in Fig. 15 the focus changed to capture whether a lie was found in a statement or not. Therefore, the ROC curve lies vs other classes and is depicted with an accuracy of 75%. ROC Micro-Average one-vs-rest is used in a multi-class classification where there are highly imbalanced classes. The results for the ROC using micro-averaging showed 86% accuracy given area under the ROC curve of Fig. 15.

TABLE I
RESULTS OF APPLYING DIFFERENT MODELS ON BALANCED DATASET WITH MULTICLASS CLASSIFICATION

| models | Accuracy | Training Accuracy | Kappa Score | F-1 | Precision | Recall | ROC Accuracy |
| --- | --- | --- | --- | --- | --- | --- | --- |
| BERT+SVM | 36.87% | 47.82% | 05.31% | 51.96% | 60.26% | 47.82% | 56.24% |
| BERT+KNN | 32.56% | 55.71% | 33.57% | 37.65% | 60.15% | 32.56% | 53.32% |
| Doc2VecC+SVM | 30.27% | 68.10% | 52.15% | 33.17% | 56.40% | 30.27% | 45.38% |
| Doc2VecC+KNN | 44.66% | 55.47% | 33.21% | 48.37% | 53.31% | 44.66% | 44.63% |
| SBERT+SVM | 45.61% | 89.40% | 84.10% | 51.17% | 65.87% | 45.61% | 63.43% |
| SBERT+KNN | 39.13% | 60.65% | 40.97% | 45.01% | 63.87% | 39.13% | 59.70% |

In classifying using KNN, the same curves as SVM are shown in Fig. 16 and 18. Fig. 18 shows the ROC curve of true vs the rest with 62% accuracy. Fig. 18 shows the ROC curve of lie vs the rest with 69% accuracy. Fig. 16 shows the ROC curve using micro-averaging with 84% accuracy.

For the balanced dataset, macro-averaging is more helpful. The results for lie vs the rest showed more accuracy like the SVM and KNN applied on the imbalanced dataset.

Different ROC curves for SVM and KNN models using an imbalanced dataset are shown in Fig. 15 and Fig. 16

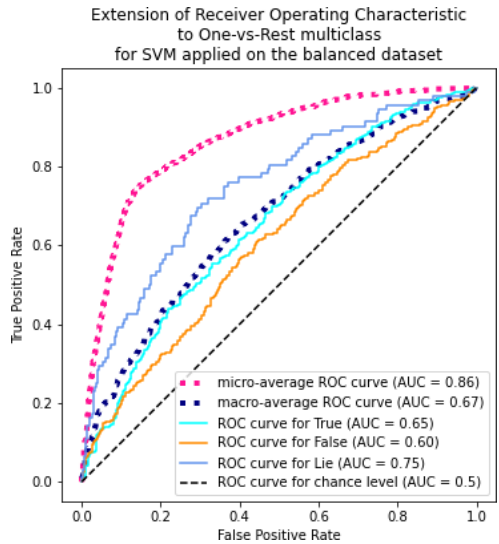

Fig. 15. ROC of SVM with Imbalanced Dataset

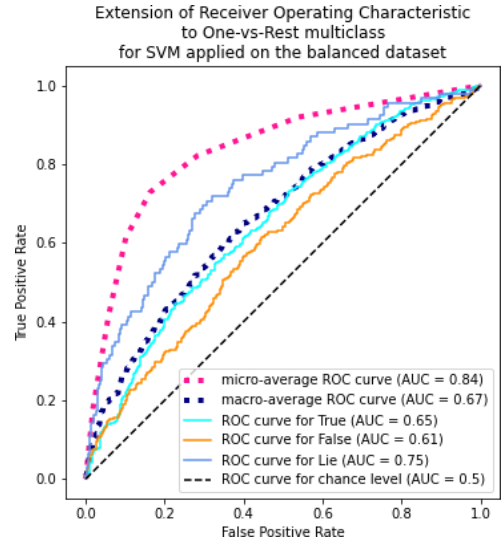

Fig. 16. ROC of KNN with Imbalanced Dataset

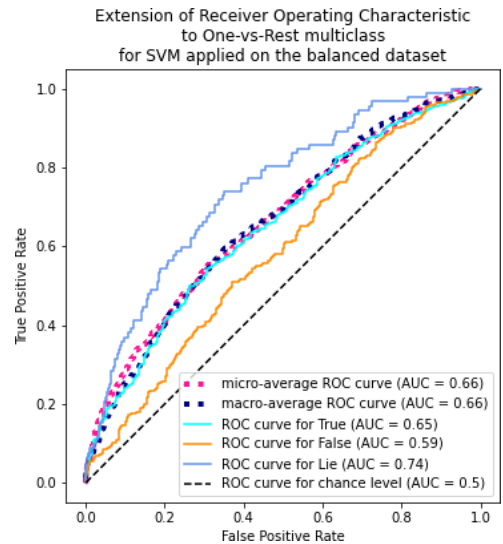

Fig. 17. ROC of SVM applied on the balanced Dataset

Since the results of the SVM and KNN classification using the balanced dataset were lower than the imbalanced dataset, two figures were plotted, where all the possible ROC curves have been shown. Fig. 17 and Fig. 18 show less area under the curve for SVM and KNN that were trained on the balanced dataset.

For the experiments on ChatGPT versus human-generated text, the model produced a perfect classification score of 1.0 for evaluation parameters and could correctly classify all 3000 text instances. To understand which words had a greater impact in deciding the classification of texts, attribution scores generated by the Integrated Gradients method as shown in Fig. 19 were used. On visualizing the top 50 score values, it was observed that the words 'reliably', 'versa', and 'era' had the highest occurrence in text generated by ChatGPT. To make the class prediction more easily understandable, the significant words were highlighted using red and green colors along with their true and predicted class labels as shown in Fig. 20. The intensity of the green shade shows how positively significant the word was in deciding the class while words in white are neutral with no contribution towards the prediction.

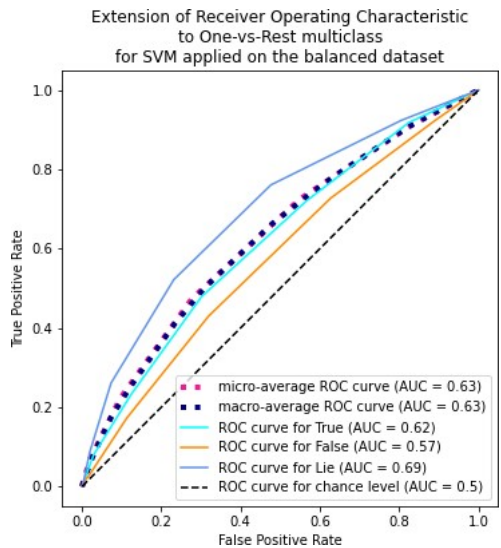

Fig. 18. ROC of KNN applied on the balanced Dataset

In the case of artificially generated and real scientific abstracts, the SVM model performed at an accuracy of 84% using sentence embeddings to classify the text. Converting the words to vectors using the CountVectorizer method and passing them through a logistic regression model resulted in an even better accuracy of 90%. The model interpretability was enhanced using LIME and SHAP based post-hoc explanations. For a specific abstract instance, the significant words for classification in-text were highlighted, and the relevance score for these words in classifying the text as fake or original was also determined.

The SHAP model provides a list of significant words across the whole corpus of text, and identifies the words 'paper', 'into', 'been', 'its', and 'such' as being most important for classifying an abstract as fake, while the words 'study', 'novel', 'levels', 'control' were most prominent in original abstracts. This could indicate that the artificially generated instances had more filler words such as the prepositions highlighted.

The use of autoencoders to visualize the text data did not yield any significant results as the points seemed to be randomly scattered along the two component axes, and no distinguishing patterns in the distribution for the two classes could be identified. There is considerable overlapping between the points for the two classes around the center of the plot, indicating similar feature representations for them. Most of the real abstracts are distributed across the top left of the plot while the fake abstracts are more spread towards the right.

The Kernel Density Estimate (KDE) plot generated shown in Fig. 21 also shows overlapping distributions for the two classes, indicating large regions where the two classes are not distinctly separated. For the human-written texts, the distribution is slightly wider, implying greater variability. The fake text distribution has a sharper and higher peak, with comparably less variability.

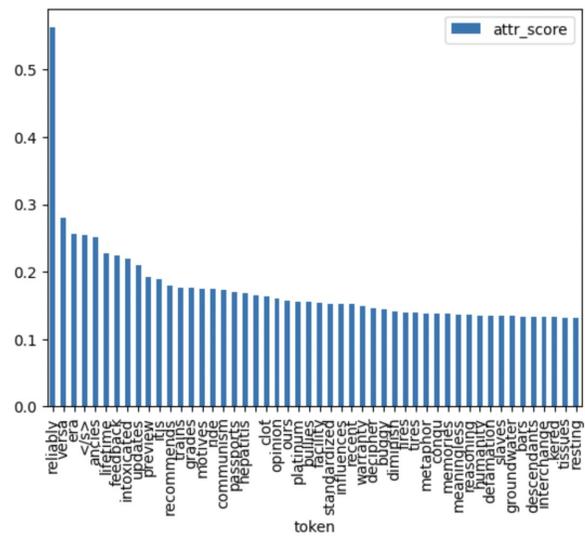

Fig. 19. Attribution scores using Integrated Gradients for top 50 words in ChatGPT generated sentences

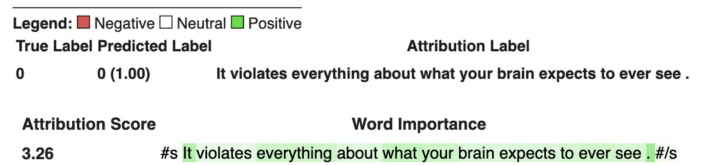

Fig. 20. Visualization of attribution scores in-text

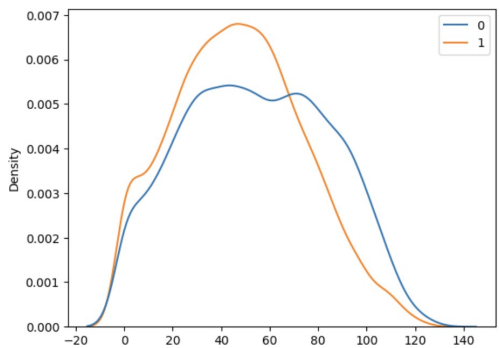

Fig. 21. Kernel Density Estimate Plot for Class Distribution

## VII. Discussion

As shown in the spectral analysis and visualization with dimensionality reduced to the two most important latent features in Figures 1, 2, 3 4, and 5, 6 and three latent features in Fig. 7 and Fig. 8, there are no natural discernible clusters of the various classes. Each class of statements is spread across the feature space and the classes are not significantly distinguishable. None of the diverse embedding schemes used place the data points in natural clusters. Instead, they place the statement vectors from different classes next to or even coinciding with each other so much so that even classifiers like SVM with RBF kernel and K-NN that can classify highly nonlinear data also cannot perform well. The numbers in Table I are indicative of this fact.

It must be noted that the visualizations are highly scaled-down versions of the original feature space but are quite representative of the relative positions of the feature vectors. Because of the compacted feature space, many feature vectors overlap with each other making it appear as if there are clusters in some combinations of the embedding scheme and spectral method. Adding a dimension and visualizing the embeddings in 3D as in Fig. 7 and Fig. 8 does not help much. However, correlating such figures with the other figures, the dataset characteristics, and also the accuracy metrics in Table I, it can be concluded that in terms of the embeddings computed, there is not a substantial difference in the embeddings for the true and other classes of textual information.

Using the LIME, SHAP, and Integrated Gradients explainability modules to improve the interpretability of the model's results for the other two sets of experiments also generated similar results. Based on the top contributing factors, in this case, words, returned by the algorithm, there are no definitive words that characterize fake and real text. Most words are generic in nature and could have been part of the corpus for either classes of text. The use of variational autoencoders to visualize the distribution of fake versus real text samples also did not yield any clear patterns for differentiating between the two classes.

This is true across the diverse embedding schemes that we used. The plots, which are of different shapes and distributions, indicate that the embeddings generated by the three frameworks are substantially different. Nevertheless, irrespective of the embedding scheme, the spectral analysis and visualization using two diverse techniques confirms our conclusion that the current representation learning approaches are unable to capture the varying degrees of truthfulness in textual information. Machine learning is primarily learning by similarity [32] [33]. Learning happens by discovering similarities and dissimilarities among data. In the case of misinformation detection, representation learning is unable to discern among similarities and dissimilarities of the varying degrees of truth in the textual information.

## VIII. Conclusion

The complexity of classifying truth from lies is detailed in this work in an attempt to answer why the problem of misinformation containment is unsolved. The work used multiple schemes of embeddings, datasets, spectral and non-spectral analysis, multiple supervised learning algorithms, explainability techniques, and evaluation metrics to analyze the reasons. Current representation learning does not significantly distinguish between true, false, and the "pants on fire" kind of information with varying degrees of truthfulness. The post-hoc explanations generated using LIME, SHAP, and Integrated Gradients do not convincingly bring out the distinction between synthetic, false, and genuine information either. A possible research direction is to invent a new embedding scheme for text that is capable of capturing the veracity of its information content.

## Acknowledgment

The authors would like to thank Foroozan Sadat Akhavan Tabatabaii for her contribution to the earlier work [5] and Rakshit Gupta for contributing to some of the experiments.